# Cooperation between Pronoun and Reference Resolution for Unrestricted Texts

**Andrei Popescu-Belis & Isabelle Robba**

LIMSI - CNRS, BP. 133

91403 — ORSAY CEDEX, FRANCE

popescu@limsi.fr — robba@limsi.fr

## Abstract

Anaphora resolution is envisaged in this paper as part of the reference resolution process. A general open architecture is proposed, which can be particularized and configured in order to simulate some classic anaphora resolution methods. With the aim of improving pronoun resolution, the system takes advantage of elementary cues about characters of the text, which are represented through a particular data structure. In its most robust configuration, the system uses only a general lexicon, a local morpho-syntactic parser and a dictionary of synonyms. A short comparative corpus analysis shows that narrative texts are the most suitable for testing such a system.

## 1 Methods for Anaphora Resolution

### 1.1 Knowledge sources

Correct interpretation of anaphora is crucial for natural language understanding systems, as it enables a system to keep track of the entities introduced through the processed text. Various knowledge sources have been used for anaphora resolution, leading to more or less realistic systems. For instance, (Hobbs, 1978) uses a parse-tree analysis algorithm, and correctly solves an average of 88% of the personal pronouns, in a selection of English texts. A blackboard-like architecture is proposed by (Rich and Luperfoy, 1988) in order to integrate various knowledge sources, but no evaluation is given.

The method proposed in (Lappin and Leass, 1994) uses context modelling and salience values, besides syntactic constraints, and proves 4% more accurate than Hobbs' algorithm on the same corpus. Salience can realistically be calculated even for unrestricted texts, and permits also integration of heterogeneous criteria. Local semantic constraints can be added to this algorithm, as in (Huls et al., 1995).

Whereas it is almost certain that complex semantic and pragmatic knowledge is needed to solve *all* the well-formed anaphors, it is highly improbable that this would soon be available for a computational system. Even elaborated semantics and complete parse trees aren't yet realistic for unrestricted text processing. A solution is then to use statistical methods to induce semantic constraints of frequently used verbs, as in (Dagan and Ito, 1990). But (Kennedy and Boguraev, 1996a) show that the Lappin and Leass algorithm still provides good results (75%) even without complete parse. They suggest also (Kennedy and Boguraev, 1996b) that anaphora resolution is part of the discourse referents resolution. However, little is said about concrete methods for building "coreference classes": the example given by the authors concerns only coreference between an acronym and its expanded form.

We describe here an open architecture for reference resolution, which provides a common frame for pronoun and reference resolution. At its most elementary level, our system uses simple cues for pronominal anaphora solving (morphology, local syntax and context rules) and simultaneously performs noun phrase referent resolution (using identity, synonyms and hyperonyms). These two aspects of the same task benefit from their cooperation.

### 1.2 The antecedent/anaphor paradigm

The cooperative strategy proposed here has long been masked by the classic conception of anaphora as a pure textual relationship, between an *anaphor* (e.g., a pronoun) and its explicit or inferred *antecedent* in the text (e.g., a noun phrase). In this view, the anaphor always needs another textual description or phrase in order to be solved, while the antecedent can refer directly to an object outside the text.

Recent work tends to unify these two situations (Ariel, 1990 and 1994) (Reboul, personal communication). They propose a gradual classification for all the *referring expressions* (RE), ranging from proper names and definite or indefinite noun phrases up to the pronouns. Their "resolution" means the construction of a link between the RE (be it nominal or pronominal) and its correct referent, from an evolving set of potential referent representations.

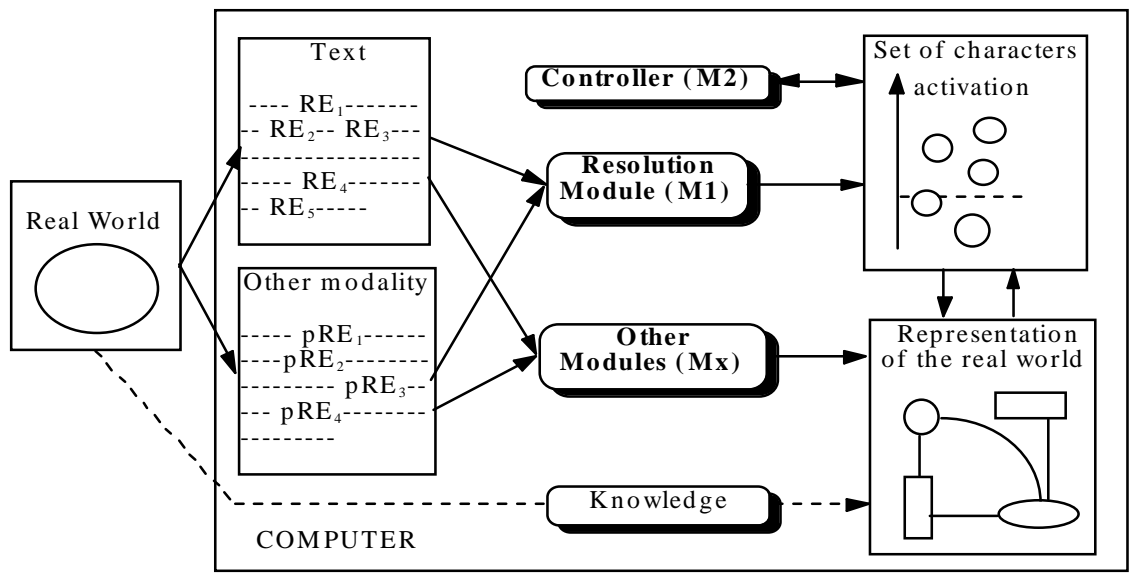

Figure 1. General structure of a reference resolution system

Therefore, we will avoid using the anaphor/antecedent distinction, and will speak instead about REs and their referent, called "character". As our open architecture supports the transition between the two paradigms, we will sometimes use also the classic terminology.

## 2 A Frame for Reference Resolution

### 2.1 General Description

We suggest that most natural language understanding systems are structured (at least partly) as in Figure 1. The machine receives natural language input (text) with referring expressions (RE), and possibly other input (e.g. mouse clicks on a screen) with pseudo-RE (pRE). Also, knowledge can be provided more directly by the programmer. The machine handles a set of referents extracted from the text - in fact *representations* of real entities, called here *characters*. A formal representation of the real world (model) may also be available.

The task of a RE resolution system is thus to build and manage a set of characters; modules M1 and M2 are its two main components. This architecture can account for:

- mono- or multi-modal interaction;
- "cognitive" system or not – depending on the model of the world;
- "classic" system or not – classic if the set of characters is just a duplicate of some of the text's phrases, not classic if an elaborate character structure is present.

### 2.2 Balance between alternatives leads to various resolution paradigms

Module M1 selects referring expressions (REs) from input data, and associates them to the proper "character structure" in the character set (cf. §3). M1 has two alternatives when solving a RE:
- (1) associate the RE to an existing character, adding new data to the character's record; – or
- (2) create a new character, its parameters being instanciated with data from the RE.

Choice restrictions can simulate various approaches. If noun phrases always call (2), and pronouns call (1), we obtain the classic antecedent/anaphora framework. Otherwise, if all categories of RE can be followed by (1) or (2), then the system treats all REs in a homogeneous way, which is cognitively more accurate.

Module M2 controls the character set, updating their activation (or salience, cf. §3.1) among other parameters. M2 can:
- (3) merge two characters in the set, if M1 has been overproductive; – or
- (4) remove (and possibly archive) characters which haven't been referred to for a long time.

At this stage, it might seem that (1) and (3) are equivalent, i.e. (2)+(3)=(1). In fact both operations are necessary as the system is given increasing knowledge. Suppressing (1) would only mean to postpone the reference resolution and leave it entirely for M2; the role of M1 would thus become insignificant. On the contrary, M1 has to start working on reference resolution, and not rely entirely on M2.

But M2 should also be able to merge two characters of the set. Indeed, before reaching a suitable balance between creating vs. merging characters ((1) vs. (2)), which is our long term goal, it is better to have an overproductive M1, privileging (1). The system would avoid incorrect resolution, which is hard to undo, and reference resolution would be at worst incomplete, but not wrong. M2 can complete the resolution by merging characters, which is much easier than undoing previous reference links. The more accurate M1 becomes, creating less and less redundant characters, the more seldom (3) is used.

The problem of revisable choices subsists however in M1, depending on how consecutive REs are treated. It is reasonable for the beginning to process the REs sequentially, validating each reference resolution before examining the next one. This rigid order is not really compatible with cataphors (unless the module can take the initiative to create a character corresponding to a pronominal RE), but has proved successful in most of the algorithms cited above. Also, it limits influence between textual close references. A better solution is to handle a buffer for the current sentence, compute mutual influence of the REs through their respective activation (cf. §5.3) and after stabilization validate the resolution for the entire current sentence. Afterwards, only M2 can make changes, by merging characters.

## 3  The character set

A character is any object, animated or inanimate, which occurs in the text: *a tree*, *a kitchen*, *a bed* may be a character. But we represent neither the events (*his marriage, the storm...*) nor the concepts of abstract domain (*a new idea, this music...*) as characters. Nevertheless, we are aware that, in a complete system, these should also be represented. We use the term "character" to refer to "the representation of a character".

### 3.1  The character structure

The structure we have adopted to describe the characters has first been proposed in (Berthelin, 1979). The main contributions of this work lie in the originality of the structure itself, and in the use of this structure to highlight the inconsistency that may be underlying in a story. This work has been primary applied to stories but its application to dialogues would not pose any problem.

The structure is involved in two processes. First, during the parsing of the text, a representational structure is instantiated for each character of the story. This structure gathers all the information about the character and underlines the different points of view of the different characters involved in the story. Second, this structure is used to detect the contradictions and the incoherence that may exist between the different points of view of the characters. Space lacks to describe the work achieved in this second step, but we will show how it improves our preliminary work.

The representation of a character C consists of a set of facets. Each of these facets contains the set of statements which have been expressed by a character C' about the character C. In that way, the facet reflects the point of view that C' has about C. Each of the statements contained in a facet consists itself of a data set: temporal references, state of C'... This information, extracted from the text, will be useful to the second aspect above.

In our approach, reference is solved without having completely parsed the text neither syntactically, nor semantically. This kind of approach is essential since we do not have actually at our disposal a parser capable of dealing with unrestricted texts. Certainly in this case, the "character" structure may appear too complex (without complete parse, there isn't enough knowledge to fill in all the structure's attributes); but the architecture is open to semantic methods which could take advantage of this complexity.

Each character **C** is described with the following parameters:
- **a label**: a number which allows to identify C
- **a list of identifiers**: the REs which have been used to design the character; we envisage to order this list according to the frequency of use of the identifiers
- **a list of verbal descriptions** (VD): what has been said about C
- **an activation value**: it represents the salience of C, this value is modified during the resolution and it depends on the context
- **an accessibility mark**: at each step of the resolution it indicates whether C is accessible or not according to the concordance rules implemented by M1 (see §2)

And each verbal description **VD** consists of:
- **a list of words**: the words which compose VD
- **a sentence number**: to localize the place of VD in the text
- **a position in the sentence**: a pair of marks which localize the RE referring to the character in the sentence

- **a f-structure**: it describes the syntactic structure of VD, if its parsing has succeeded

The values of the different parameters of the structure are determined by the module M1, except for the f-structure which is not always available.

## 3.2 Modifications of the character set

The reference resolution mechanism consists in the interaction of two modules (namely M1 and M2, cf. §2). M2 periodically examines the complete set of characters (provided by M1), to determine whether two or more characters should be merged into a single one. Indeed, since our system does not dispose of all the knowledge necessary to understand correctly a text it may make mistakes which a merging module might be able to rectify on the basis of further information.

Moreover, a complete system of text comprehension should be able to dynamically modify the set of characters. Indeed, even when parsing the best written texts, comprehension mechanisms sometimes have to backtrack on their decisions, and on the characters they have recognised.

Obviously, a module able to detect the inconsistencies (the one proposed by Berthelin) would be essential to give some indications for launching the merging module, but since it is not actually available, we suggest to trigger it with a regular but arbitrary frequency. We also suggest a mechanism which tracks M1, and triggers the merging when several characters have been created due to the presence of definite determinants, because this kind of determinant often describes a character already introduced, but without using the terms already used.

As far as the structure is concerned, the mechanism has to unify the parameters of the two characters. The following methods are proposed:

- **the label**: only the smaller of the two numerical numbers is retained (the label of the first RE which introduced the character)
- **the list of identifiers**: the 2 lists are merged
- **the list of verbal descriptions**: the two lists are simply merged, and the parameters constituting the VD are conserved
- **the activation value**: it seems reasonable to retain the higher of them, but a more complex calculation may also be considered
- **the accessibility mark**: there is no decision to take since this mark is determined at each resolution.

## 4 Processing French Texts

### 4.1 Description of three corpora

The system presented here is designed to work on unrestricted French texts; therefore, only few robust NLP resources are available. Non-specialized texts are preferable, as convenient lexicons are available for general vocabulary. We considered three texts: an essay by Stendhal (from the *Chroniques Italiennes*), a scientific report by Gérard Sabah (PLC), and some Stock Market articles from the journal *Le Monde*. The next table compares characteristics of these texts, and indicates that Stendhal is the most rich and interesting from the anaphora point of view.

|  | Stendhal | PLC | Le Monde |
|---|---|---|---|
| Word count | 9 144 | 15 006 | 16 504 |
| /il/ personal (he,it) | 165 | 50 | 35 |
| /il/ impersonal (~it) | 44 | 45 | 45 |
| /elle/ (she,it) | 31 | 19 | 25 |
| /le/ masc. art. (the) | 217 | 226 | 367 |
| /le/ masc.pron.(him,it) | 16 | 7 | 3 |
| /la/ fem. art. (the) | 201 | 345 | 577 |
| /la/ fem. pron. (her,it) | 7 | 2 | 1 |
| /l'/ article | 95 | 374 | 242 |
| /l'/ masc. pron. | 13 | 6 | 2 |
| /l'/ fem. pron. | 7 | 4 | 0 |
| /lui/ masc. indirect obj. | 37 | 5 | 3 |
| /lui/ fem. indirect obj. | 10 | 1 | 5 |
| /lui/ masc. tonic pron. | 20 | 2 | 6 |
| /son/, /sa/, /ses/ poss. | 110 | 46 | 87 |

### 4.2 Ambiguity of French pronouns

Notwithstanding our critique of the antecedent/anaphor distinction, we focus for the beginning on pronoun resolution. We examine the 3rd person singular and plural, subject, direct object and indirect object pronouns: /il/, /elle/, /le/, /la/, /l'/, /lui/, /ils/, /elles/ (the English /he/, /she/, /him/, /her/, /it/, /they/).

Three main problems appear specific to French. First, /le/ and /la/ are both pronouns and definite articles, so one has to select pronominal occurrences before the reference resolution. Second, elision and use of an apostrophe for /le/ and /la/ change them into the even more ambiguous form /l'/, which has four interpretations; as a pronoun, all indication of gender disappears. Third,

/lui/ can be an indirect object pronoun, masculine and feminine, and also the tonic form of /il/.

## 5 Realisation

### 5.1 General overview

Module M1 selects nominal or pronominal REs in the input text. We impose that new pronominal REs be always linked to existing characters, M1=>(1), as detailed at the end of §2.2. There is clearly a need for first instanciating the character set ("antecedents", in the classical terminology): M1 processes also nominal REs (noun-phrases) from the text. When processing a NP–RE, M1 can choose between (1) or (2), i.e. create a new character (like in classic systems) or link the RE to an existing character.

The resulting mechanism is now easy to understand. M1 reads linearly the (pre-processed) input text, and when it finds a NP–RE, either attaches it to a previous character if the linguistic descriptions match (same word, synonym, hyperonym), or builds a new character with the corresponding description and activation. When M1 processes a pronominal RE, it uses "salience value" criteria (cf. (Lappin and Leass, 1994) and (Huls, 1995)), intertwined with morpho-syntaxic constraints (and later semantic ones), in order to choose a character from the set as referent of the RE. The character's parameters are then updated, in particular its linguistic descriptions and activation.

### 5.2 Resources used by the system

Robust linguistic resources are essential for processing unrestricted texts. The most important one is an LFG parser developed in the Language and Cognition Group at the LIMSI (Vapillon et al., 1997); however, as the rules cannot yet cover a significant proportion of complex sentences, our system uses only local analysis, which parses NPs even when the sentence analysis fails. Thus, the only limitations are the lexicon used by the parser[1], some complex NPs, and, of course, morpho-syntactic ambiguities.

A tagger is used to help lexical disambiguation, and performs also robust pre-processing of the input text. The STK tagger (Ferrari, 1996) developed at the LIMSI is used together with some simple rules for distinguishing the article /le/, /la/ from the pronoun /le/, /la/. When these rules aren't sufficient (e.g., unknown or truly ambiguous noun), we don't consider /le/, /la/ as an article.

---

[1] The lexical analyser has a dictionary of 25 000 canonical forms and 350 000 inflected forms.

Our proposition for a robust reference resolution relies on two ideas. First, a character is often designated by the same phrase – this criterion is extremely simple to track, and is worth considering. Second, an entity is often designated by a synonym of the previously used RE, or a hyperonym. It is thus interesting to use a dictionary of synonyms, and we are currently integrating one.

### 5.3 Activation values and selection

Activation is the global salience value of each character. Several gradual criteria are used: the more a criterion is satisfied by a character, the higher its contribution to the character's activation will be. The following criteria have been implemented:
- recency of the last textual mention of the character (last RE)
- number of REs already referring to the character: mention by a nominal RE brings more activation than a pronominal RE, and proper NPs bring more activation than common NPs
- grammatical role of the last textual mention (last RE). Activation decreases from subject to direct object, indirect object, or other.

Behind the elegance of the activation paradigm, which integrates different criteria (and possibly multi-modality), there is a hidden limitation. The activation distribution does not depend on the nature of the current RE, but only on its position. So, character activation cannot take into account properties of the processed RE, and syntactic parallelism cannot be considered, as it would require the activation to depend on the RE's nature.

Besides, the activation distribution at a given point in the text is a recursive function: it is calculated using also its previous values. This makes backtracking (and revisable choices) difficult to implement, as they would require a complete recomputation of the activation distribution (or a "decomputation"). That is why M1's choices aren't revisable for the moment.

Finally, the system has to take somewhere into account the RE's nature, and operate a selection among the characters. This is done without further computation, using a set of various constraints which change the binary value (yes/no) of the character's "accessibility mark". In this way, only "accessible" characters are considered when solving a particular RE, and all "accessibility marks" are subsequently reset to "true".

The selectional constraints implemented at this stage are:
- for NP-REs, gender and number concordance
- for pronominal REs, number and gender, if unambiguous (cf. the /l'/, § 5.3). Furthermore,

- an object pronoun cannot refer to the subject of the sentence
- coreference is hypothesized if two NP-REs are identical, or if the second is a hyperonym of the first.

## 6 Results

The system is implemented in Smalltalk, and its user-friendly interface permits step-by-step monitoring of the process as well as parameter tuning.

Current work concerns Stendhal's text, as it has the highest density and variety of pronouns (cf. § 4.1). Its syntactic complexity and the overproductivity of the local LFG parser oblige us to make manual selection among NPs, and disambiguation of the /le/, /la/, /l'/ pronouns vs. definite articles. The next table summarizes our first experiment.

| DATA | |
|---|---|
| Words | 3954 |
| Sentences | 131 |
| Nominal REs | 495 |
| Pronominal REs | 113 |
| REs per sentence | 3.8 |
| **RESULTS** | |
| Characters found | 291 |
| Pronouns **correctly** attached | 70 (62%) |

When coreference is not dealt with, there are as many "characters" as NPs, and, as expected, the number of correctly solved pronouns is smaller (by 40%).

The pronoun resolution score, 62%, is a little smaller than those obtained elsewhere for English texts; but these results are encouraging, especially as they rely only on simple rules. Moreover, on this particular text, we have observed, that 50% of the mistakes could be avoided using on the one hand simple semantic constraints derived from verbal argument structure (e.g., human/non-human subject, animated/non-animated subject/object…); on the other hand syntactic constraints concerning the possibilities of coreference between NPs and pronouns occurring in the same sentence.

Further work will first concern a more accurate tuning of the parameters, and adjunction of new activation and selection rules. In particular, syntactic restrictions will be adapted to the local parser's data. Also, we would like to make the processing entirely automatic, which requires a selection among the NPs provided by the local parser, and disambiguation of the pronouns. These being complex tasks, they will probably decrease the success rate, especially with respect to English, where articles and pronouns are never homonymous.

## Acknowledgements

The authors are grateful to Anne Reboul, Gérard Sabah and Jean-Baptiste Berthelin for valuable discussions and advice. This work is related to the CERVICAL research project, involving members of the LIMSI (Orsay) and CRIN (Nancy).